%% file: hypermcts.tex
\documentclass{article}
\usepackage{preprint,times}
\input{math_commands.tex}

\usepackage[hidelinks]{hyperref}
\usepackage{url}
\usepackage{amsmath,amssymb,amsfonts}
\usepackage{algorithmic}
\usepackage{graphicx}
\usepackage{subcaption}
\usepackage{textcomp}
\usepackage[T1,OT1]{fontenc}
\usepackage{xcolor}
\usepackage{algorithm}
\usepackage{booktabs}
\usepackage{pifont}
\usepackage{multirow}
\usepackage{rotating}
\usepackage{placeins}
\newcommand{\tablehead}[1]{\begin{tabular}[c]{@{}c@{}}#1\end{tabular}}
\makeatletter
\def\fps@figure{!t}
\def\fps@table{!t}
\makeatother

\title{HyperMCTS: Hypergraph-Augmented MCTS \\ for Long-Horizon LLM Agents}
\author{%
\makebox[\textwidth][l]{%
\begin{minipage}{\textwidth}
\raggedright
\textbf{Tingsong Xiao}\textsuperscript{1}\hspace{1em}
\textbf{Nithish Balachandar Moudhgalya}\textsuperscript{2}\hspace{1em}
\textbf{Chandrayee Basu}\textsuperscript{2}\\[0.1em]
\textbf{Lichao Wang}\textsuperscript{2}\hspace{1em}
\textbf{Luyang Kong}\textsuperscript{2}\hspace{1em}
\textbf{Benjamin Z. Yao}\textsuperscript{2}\hspace{1em}
\textbf{Zhe Jiang}\textsuperscript{1}\hspace{1em}
\textbf{Jie Hao}\textsuperscript{2}\\[0.2em]
\textsuperscript{1}University of Florida
\hspace{1em}
\textsuperscript{2}Amazon\\[-0.1em]
{\small\texttt{xiaotingsong@ufl.edu, jieha@amazon.com}}
\end{minipage}%
}%
}

\date{}
\hypersetup{pdftitle={HyperMCTS: Hypergraph-Augmented MCTS for Long-Horizon LLM Agents},pdfauthor={Tingsong Xiao, Nithish Balachandar Moudhgalya, Chandrayee Basu, Lichao Wang, Luyang Kong, Benjamin Z. Yao, Zhe Jiang, Jie Hao},pdfsubject={Preprint},pdfkeywords={Language model agents, Monte Carlo tree search, hypergraphs, planning}}
\begin{document}
\maketitle

\begin{abstract}
Long-horizon tasks require large language model (LLM) agents to coordinate decisions under constraints that span an entire solution. Monte Carlo Tree Search (MCTS) offers a promising approach to test-time scaling by exploring alternative action trajectories, but model computation and environment interaction make search costly. Efficient search therefore requires effective reuse of trajectory feedback. Standard MCTS maintains prefix-specific statistics, without explicitly accumulating outcomes for decision groups that recur across different paths. To fill this gap, we propose \textbf{HyperMCTS}, a training-free method that augments an ordered MCTS tree with a cross-trajectory hypergraph. Hyperedges represent groups of canonical decisions and accumulate their observed returns within the current task. Our hypergraph-guided \textbf{HyperUCT} selection rule aggregates evidence from overlapping hyperedges into an action prior, allowing outcomes collected under one prefix to inform selection under another while preserving execution histories in the tree. On DeepPlanning, HyperMCTS improves average planning accuracy by 2.3--7.3 percentage points over the strongest baseline for each of three backbone models. It enables Qwen3.6-27B to outperform Claude Opus 4.6 (max) on Shopping Planning, while achieving higher accuracy with fewer LLM calls and output tokens than the evaluated MCTS-based baselines. SealQA experiments further demonstrate improvements in question answering.
\end{abstract}

\section{Introduction}
\label{sec:intro}

\begin{figure}[!t]
\centering
\includegraphics[width=0.98\linewidth]{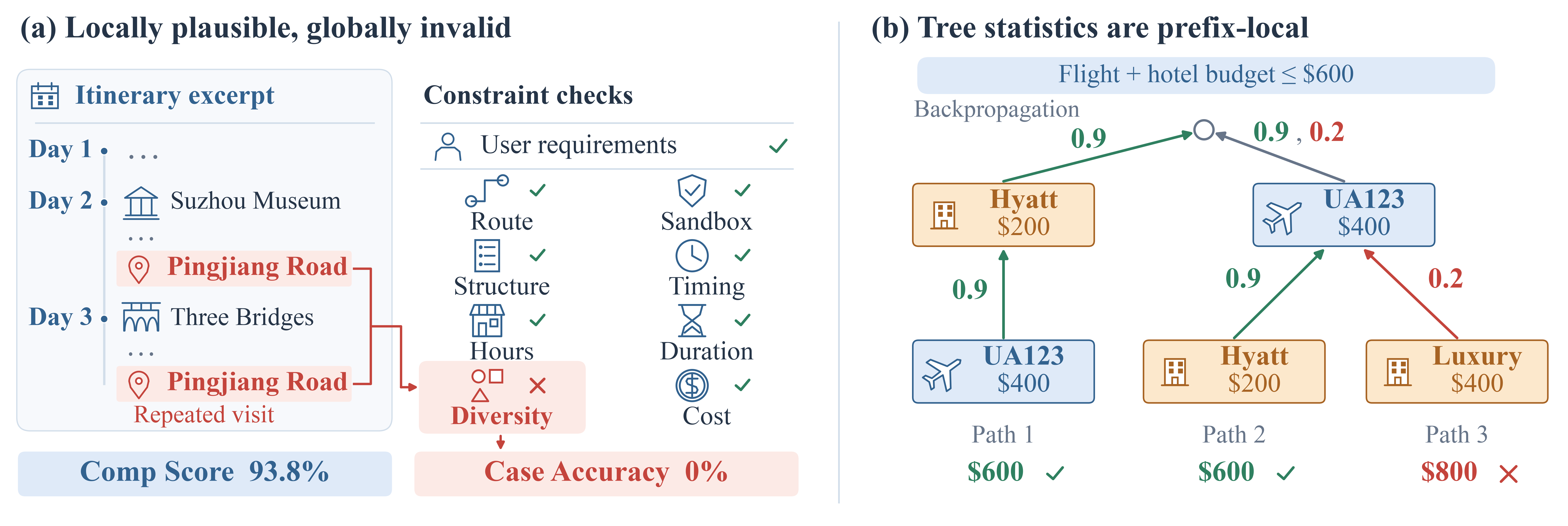}
\caption{(a) A recorded DeepPlanning Travel failure despite 93.8\% Comp Score. (b) An independent schematic under a \$600 budget: Paths 1 and 2 construct the same flight--hotel combination in different orders; Path 3 changes the hotel. Returns 0.9 and 0.2 are illustrative feedback, not measurements or values derived from the budget check.}
\label{fig:motivation}
\end{figure}

Large language model (LLM) agents are increasingly studied for long-horizon tasks such as web navigation, software engineering, and travel or shopping planning~\citep{zhou2024webarena,antoniades2025swesearch,zhang2026deepplanning}. These tasks require tool interactions and decisions whose consequences extend across the complete solution: individually plausible choices may jointly violate a budget or conflict in time. Figure~\ref{fig:motivation}(a) illustrates this difficulty: a DeepPlanning itinerary attains a Composite Score of 93.8\% but fails because an attraction is scheduled twice.

Test-time scaling (TTS) allocates additional inference computation to improve task solving~\citep{snell2024scaling,zhang2025ttssurvey}. Chain-of-thought encourages intermediate reasoning~\citep{wei2022chain}, and self-consistency aggregates answers from sampled reasoning paths~\citep{wang2023selfconsistency}. For interactive agents, ReAct interleaves reasoning and interaction~\citep{yao2023react}, while Reflexion revises attempts using feedback~\citep{shinn2023reflexion}. These approaches support deliberation and revision without explicitly maintaining a search tree of alternative action histories.

Structured methods organize reasoning into thought trees or graphs~\citep{yao2023tree,besta2024graph}. HyperTree Planning~\citep{gui2025hypertree} uses hyperedges for hierarchical subtask decomposition, but guides search through LLM judgments rather than value estimates updated from rollout returns. Monte Carlo Tree Search (MCTS) evaluates candidate actions through simulated continuations and backs up their returns to guide subsequent search, balancing exploration with exploitation~\citep{kocsis2006bandit,browne2012survey}. Applications include world-model reasoning in RAP~\citep{hao2023reasoning}, self-refinement in MCTSr~\citep{zhang2024accessing}, trajectory verification in rStar~\citep{qi2024mutual}, and value-guided decoding in PPO-MCTS~\citep{liu2024value}. For interactive agents, LATS incorporates environment feedback and reflection~\citep{zhou2024language}; ExACT demonstrates search-budget scaling for long-horizon web agents~\citep{yu2025exact}; SWE-Search explores inference-time scaling for software engineering~\citep{antoniades2025swesearch}; and FLARE combines lookahead with repeated replanning~\citep{wang2026flare}. These results motivate further study of MCTS-based test-time scaling for long-horizon agents.

However, each evaluated trajectory requires model computation and environment interaction, making search costly~\citep{yu2025exact}. Effective scaling therefore depends on reusing its feedback. Standard MCTS stores values for history--action pairs, without an explicit entry for decision groups recurring across prefixes. In Figure~\ref{fig:motivation}(b), two paths construct the same flight--hotel combination in different orders, yet update separate tree paths. Another path changes the hotel and produces a different budget outcome. This motivates sharing observed outcomes at the group level, particularly when constraints couple multiple decisions.

Prior work reuses evidence through shared action values~\citep{gelly2011rave}, reasoning graphs~\citep{gong2026regrapht}, trajectory memory~\citep{wang2026flare}, and textual experience~\citep{li2026memory,xie2026sgamcts}. However, these methods do not maintain numerical outcome statistics indexed by recurring, unordered decision groups and aggregate evidence across overlapping groups to guide MCTS selection within the current task.

To fill this gap, we propose \textbf{HyperMCTS}, a training-free method augmenting ordered MCTS with a cross-trajectory hypergraph. Nodes identify canonical decisions across prefixes; hyperedges accumulate counts and mean returns for decision groups. \textbf{HyperUCT} aggregates incident hyperedge statistics into an action prior that complements tree values and exploration. The tree retains execution history while the hypergraph shares group-level evidence. This sharing is approximate because canonical identities may omit context affecting decision outcomes. We target settings where alternative histories can be replayed or simulated.

We evaluate HyperMCTS on DeepPlanning's Travel and Shopping tasks~\citep{zhang2026deepplanning}. Under the evaluated configurations, it improves average planning accuracy over each backbone model's strongest baseline by $7.3$, $2.7$, and $2.3$ percentage points for Qwen3.6-27B, Qwen3.5-27B, and Sonnet 4.6, respectively. HyperMCTS enables Qwen3.6-27B to outperform Claude Opus 4.6 (max) on Shopping Planning, reaching $60.8\%$ Case Accuracy compared with the publicly reported $56.2\%$. In the search-efficiency study, HyperMCTS also achieves higher accuracy with fewer LLM calls and output tokens than the evaluated MCTS-based baselines. Additional experiments on SealQA~\citep{pham2025sealqa} show improvements in question answering, extending the evaluation beyond planning.

Our contributions are threefold:
\begin{itemize}
    \item A training-free, hypergraph-augmented MCTS framework sharing decision-group outcomes across trajectories while preserving ordered execution histories.
    \item HyperUCT selection combining cross-trajectory priors with tree values and exploration.
    \item Evaluation of accuracy and search cost on DeepPlanning, with hypergraph ablations, weight sensitivity, and generalization to question answering on SealQA.
\end{itemize}

\section{Problem Statement}
\label{sec:problem}

\subsection{Preliminaries}
\label{subsec:preliminaries}

We consider an LLM agent interacting with an environment through tools, or application programming interfaces (APIs). Following prior work~\citep{hao2023reasoning,zhou2024language,yu2025exact}, we formulate this interaction as a finite-horizon partially observable sequential decision process. Let $\mathcal{Z}$, $\mathcal{A}$, and $\mathcal{O}$ denote the latent state, action, and observation spaces. Action $a_t\in\mathcal{A}$ changes state $z_t\in\mathcal{Z}$ according to an unknown kernel $P(z_{t+1}\mid z_t,a_t)$ and yields observation $o_{t+1}\in\mathcal{O}$. The agent acts from its observable history state
\begin{equation}
\label{eq:history}
h_t=(x,o_0,a_0,o_1,\ldots,a_{t-1},o_t),
\end{equation}
where $x$ is the user query and $o_0$ the initial observation. A frozen LLM with parameters $\theta$ defines a policy $\pi_\theta(a_t\mid h_t)$ over valid actions. An episode ends with a final answer, a task-defined failure, or the horizon $H$, producing a trajectory\footnote{We use trajectory and rollout interchangeably throughout the paper. A tree path refers to the portion of a trajectory explicitly represented in the search tree.}
\begin{equation}
\label{eq:trajectory}
\tau=(x,o_0,a_0,o_1,\ldots,a_{T-1},o_T),\qquad T\leq H.
\end{equation}

In planning, information-gathering actions $\mathcal{A}_{\mathrm{info}}$ acquire evidence, whereas commit actions $\mathcal{A}_{\mathrm{com}}$ select an item, document, or plan component. The tree branches at these decision points, retaining information-gathering steps in the history. We denote the valid search decisions at $h$ by $\mathcal{A}_{\mathrm{dec}}(h)$; their task-specific granularity is described in Appendix~\ref{appendix:adapters}. They label tree edges and map to canonical hypergraph nodes. A commit fixes part of a candidate solution, without necessarily executing a real-world transaction.

A task specifies $m$ objectives $\mathcal{G}=\{g_1,\ldots,g_m\}$, including local requirements and global constraints. A terminal verifier $V$ provides the trajectory reward
\begin{equation}
\label{eq:reward}
R(\tau)=V(\tau;\mathcal{G})\in[0,1].
\end{equation}
Here $R$ is search feedback (Appendix~\ref{appendix:reward}) and need not equal a final benchmark metric. When available, objective-level feedback $\rho_g(\tau)\in[0,1]$ supplements these statistics without replacing $R$. Global constraints, such as a total budget, depend jointly on several decisions, so satisfying most individual requirements need not yield a correct final plan. A list of commonly used notation is provided in Table~\ref{tab:notation} in Appendix~\ref{appendix:notation}.

\subsection{Problem Definition}
\label{subsec:objective}

Let $\mathfrak{T}(x)$ be the set of trajectories that are feasible under the environment dynamics, tool schemas, and horizon $H$. The ideal planning objective is
\begin{equation}
\tau^\star \in \arg\max_{\tau\in\mathfrak{T}(x)} R(\tau).
\label{eq:objective}
\end{equation}
At test time, however, the planner can evaluate only a finite set of trajectories. Given a budget $B$ of search iterations, let $\mathcal{D}_B\subset\mathfrak{T}(x)$ denote the trajectories generated and evaluated using the frozen policy $\pi_\theta$. The operational objective is therefore to return
\begin{equation}
\label{eq:budgeted-objective}
\widehat{\tau}_B \in \arg\max_{\tau\in\mathcal{D}_B} R(\tau),
\qquad |\mathcal{D}_B|\leq B,
\end{equation}
without updating $\theta$. The problem is to allocate this budget so that evidence obtained from one rollout can inform other rollouts, including those that do not share the same action prefix.

\section{Method: HyperMCTS}
\label{sec:method}

\subsection{Overview}
\label{subsec:overview}
\begin{figure}[!t]
\centering
\includegraphics[width=\linewidth]{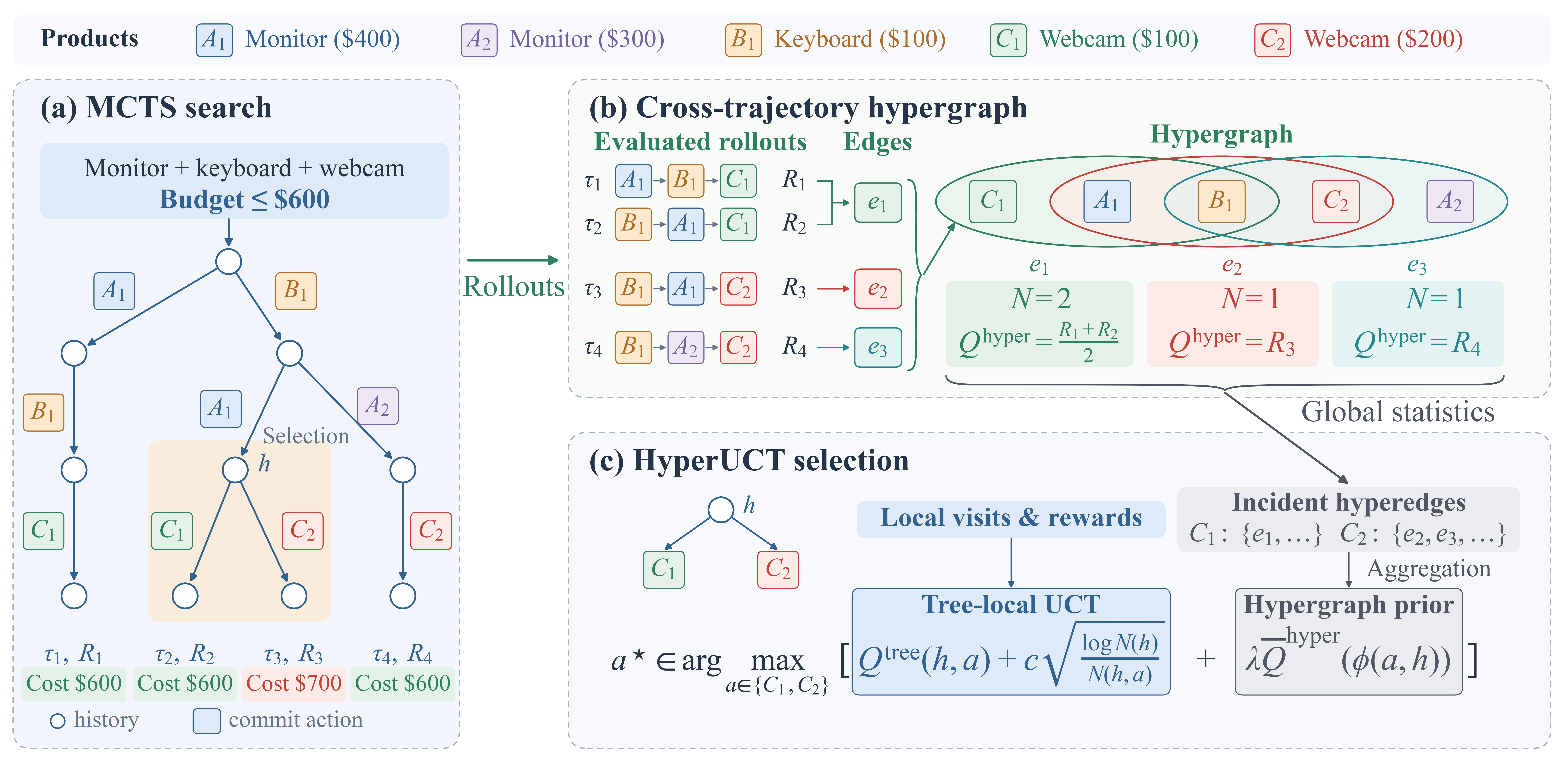}
\caption{HyperMCTS illustrated with shopping decisions. (a) Four rollouts through a tree of commit actions. (b) Shared and overlapping action groups accumulate counts and mean feedback across trajectories. (c) At the highlighted history $h$, HyperUCT combines tree-local UCT (blue) with a prior obtained by aggregating incident-hyperedge statistics (gray). $R_i$ denotes symbolic search-time feedback; costs indicate budget feasibility only. The panels show successive operations. Only trajectory-level hyperedges are drawn; search-iteration subscripts are omitted.}
\label{fig:overview}
\end{figure}

HyperMCTS augments an ordered history tree $\mathcal{T}_{\mathrm{tree}}$ with a cross-trajectory hypergraph $\mathcal{H}_k=(\mathcal{V}_k,\mathcal{E}_k)$ after $k$ evaluated rollouts of the current task. Figure~\ref{fig:overview} illustrates the method with a shopping example involving a monitor, keyboard, and webcam under a \$600 budget.

In panel (a), the same products can be selected along different tree paths, as in $A_1B_1C_1$ and $B_1A_1C_1$. Panel (b) maps these products to shared decision nodes and pools the two rollouts' feedback in hyperedge $e_1$, while overlapping product groups form other hyperedges. In panel (c), after choosing $B_1$ then $A_1$, the agent compares webcams $C_1$ and $C_2$. HyperUCT combines tree-local statistics with priors aggregated from each candidate's incident hyperedges, allowing feedback from other branches to guide this choice. Appendix~\ref{appendix:case} details the example.

\subsection{Cross-Trajectory Hypergraph}
\label{subsec:hypergraph}
A hypergraph allows one edge to connect multiple decision nodes~\citep{gallo1993directed}. For a valid decision $a\in\mathcal{A}_{\mathrm{dec}}(h)$, a domain-specific map assigns a canonical identity $v=\phi(a,h)\in\mathcal{U}$, where $\mathcal{U}$ is the space of decision identities. The map retains task-relevant distinctions while removing tree location. Let $\mathcal{C}(\tau)\subseteq\mathcal{U}$ collect these identities for the recorded search decisions in trajectory $\tau$, which may include all search decisions or a task-defined subset. The node set is $\mathcal{V}_k=\bigcup_{i=1}^{k}\mathcal{C}(\tau_i)$.

With label space $\mathcal{Y}=\mathcal{G}\cup\{\mathrm{global}\}$, each hyperedge has the key
\begin{equation}
\label{eq:hyperedge}
e=(S_e,y_e),\qquad \emptyset\neq S_e\subseteq\mathcal{V}_k,
\quad y_e\in\mathcal{Y}.
\end{equation}
Each hyperedge is indexed by its decision set $S_e$ and label $y_e$, while observed scores update its statistics. Trajectory-level hyperedges use $y_e=\mathrm{global}$; objective-level hyperedges use $y_e=g\in\mathcal{G}$ for decisions associated with objective $g$.

After a rollout, the extractor $\Psi$ returns observations
\begin{equation}
\label{eq:edge-extraction}
\Psi(\tau,\mathcal{G})=\bigl((S_j,y_j,r_j)\bigr)_{j=1}^{J_\tau},
\qquad \emptyset\neq S_j\subseteq\mathcal{C}(\tau),\quad y_j\in\mathcal{Y},\quad r_j\in[0,1].
\end{equation}
Here $J_\tau$ is the number of observations. A trajectory-level observation pairs $\mathcal{C}(\tau)$ with reward $R(\tau)$ when at least two decisions are recorded. Objective-level observations use $\rho_g(\tau)$ when available, or $R(\tau)$ otherwise. Using $R$ records an association with the whole-trajectory outcome, rather than independent objective satisfaction. Appendix~\ref{appendix:hyperimpl} specifies the extraction rules.

Starting from $\mathcal{E}_0=\emptyset$, the observations $((S_j,y_j,r_j))_{j=1}^{J_{\tau_k}}=\Psi(\tau_k,\mathcal{G})$ update the edge set as
\begin{equation}
\label{eq:edge-set-update}
\mathcal{E}_k=\mathcal{E}_{k-1}\cup
\{(S_j,y_j):1\leq j\leq J_{\tau_k}\}.
\end{equation}
For each previously unseen key $e$, initialize its count $N^{\mathrm{hyper}}(e)$ and mean $Q^{\mathrm{hyper}}(e)$ to zero. Each observation for $e$ with score $r$ updates
\begin{equation}
\label{eq:qedge}
\begin{aligned}
N^{\mathrm{hyper}}(e)&\leftarrow N^{\mathrm{hyper}}(e)+1,\\
Q^{\mathrm{hyper}}(e)&\leftarrow Q^{\mathrm{hyper}}(e)
+\frac{r-Q^{\mathrm{hyper}}(e)}{N^{\mathrm{hyper}}(e)}.
\end{aligned}
\end{equation}
Subscript $k$ denotes statistics after rollout $k$. Observations with the same key accumulate across prefixes even when their rewards differ. This pooling does not assume that the task is order-invariant; the tree retains the valid execution histories.

\subsection{HyperUCT Selection}
\label{subsec:hyperuct}
For a decision node $v$, let $\mathcal{E}_k(v)=\{e=(S_e,y_e)\in\mathcal{E}_k:v\in S_e\}$ be its incident hyperedges. Their statistics define the action-level prior
\begin{equation}
\label{eq:qhyper}
\overline{Q}_k^{\mathrm{hyper}}(v)
=\operatorname{Agg}\bigl((y_e,Q_k^{\mathrm{hyper}}(e))_{e\in\mathcal{E}_k(v)}\bigr),
\end{equation}
where $\operatorname{Agg}$ combines objective-level and trajectory-level records into a value in $[0,1]$, as specified in Appendix~\ref{appendix:hyperimpl}; the prior is zero when $\mathcal{E}_k(v)$ is empty.

Let $N_k(h)$ be the visit count of history $h$, $N_k(h,a)$ the backup count of edge $(h,a)$, and $Q_k^{\mathrm{tree}}(h,a)$ its mean backed-up reward. Among expanded valid decisions $\mathcal{A}_{\mathrm{tree}}(h)\subseteq\mathcal{A}_{\mathrm{dec}}(h)$, HyperUCT selects
\begin{equation}
\label{eq:hyperuct}
 a^\star\in\arg\max_{a\in\mathcal{A}_{\mathrm{tree}}(h)}
 \left[
 \underbrace{Q_k^{\mathrm{tree}}(h,a)}_{\text{local tree exploitation}}
 +\underbrace{c\sqrt{\frac{\log N_k(h)}{N_k(h,a)}}}_{\text{local tree exploration}}
 +\underbrace{\lambda\overline{Q}_k^{\mathrm{hyper}}(\phi(a,h))}_{\text{cross-trajectory prior}}
 \right],
\end{equation}
where $c>0$ controls exploration and $\lambda\geq0$ weights the hypergraph prior. Unvisited expanded actions receive infinite score, as in UCT~\citep{kocsis2006bandit}. Setting $\lambda=0$ recovers the UCT scoring rule.

The tree value estimates continuation reward at a specific prefix, while the prior summarizes outcomes of groups containing $v$ across trajectories. The prior does not condition on the complete current decision set and is therefore a heuristic, not a guarantee of compatibility. It guides search without changing the verifier reward $R(\tau)$.

\subsection{The HyperMCTS Algorithm}
\label{subsec:algorithm}
Algorithm~\ref{algo:hypermcts} follows the standard selection, expansion, simulation, and backup cycle, with an additional hypergraph update after each evaluated rollout. Each nonterminal node is expanded at most once, adding up to $b$ children corresponding to distinct decisions. Iteration $k$ reads statistics from the preceding $k-1$ rollouts. The candidate set $\mathcal{D}_k$ contains evaluated trajectories, as in Section~\ref{subsec:objective}, and the final choice uses their stored rewards.

\begin{algorithm}[!t]
\caption{HyperMCTS}
\label{algo:hypermcts}
\begin{algorithmic}[1]
\REQUIRE Initial history $h_0$, frozen policy $\pi_\theta$, verifier $V$, objectives $\mathcal{G}$, budget $B$, branching factor $b$, and coefficients $c,\lambda$
\ENSURE Best evaluated trajectory $\widehat{\tau}_B$
\STATE Initialize $\mathcal{T}_{\mathrm{tree}}$ at $h_0$, $\mathcal{H}_0\leftarrow(\emptyset,\emptyset)$, and $\mathcal{D}_0\leftarrow\emptyset$
\FOR{$k=1$ \TO $B$}
    \STATE \textbf{Selection:} descend from $h_0$ using HyperUCT with statistics at $k-1$ until reaching an unexpanded or terminal history $h$
    \STATE \textbf{Expansion:} if $h$ is nonterminal, sample up to $b$ distinct valid actions in $\mathcal{A}_{\mathrm{dec}}(h)$ from $\pi_\theta$, add their child histories, and set $h$ to one selected child
    \STATE \textbf{Simulation:} complete trajectory $\tau_k$ from $h$ with $\pi_\theta$ and environment feedback, or retain it if already terminal
    \STATE \textbf{Evaluation:} obtain $R_k=V(\tau_k;\mathcal{G})$ and any available objective-level feedback
    \STATE \textbf{Tree update:} back-propagate $R_k$ along the tree path from $h$ to $h_0$, including any edge selected during expansion, to update node and edge counts and $Q_k^{\mathrm{tree}}$
    \STATE \textbf{Hypergraph update:} set $\mathcal{V}_k\leftarrow\mathcal{V}_{k-1}\cup\mathcal{C}(\tau_k)$, extract $\Psi(\tau_k,\mathcal{G})$, update $\mathcal{E}_k$ by Eq.~\ref{eq:edge-set-update}, and apply Eq.~\ref{eq:qedge} to each extracted observation
    \STATE $\mathcal{D}_k\leftarrow\mathcal{D}_{k-1}\cup\{\tau_k\}$
\ENDFOR
\RETURN $\widehat{\tau}_B\in\arg\max_{\tau\in\mathcal{D}_B}R(\tau)$
\end{algorithmic}
\end{algorithm}

Rollouts terminate under the conditions in Section~\ref{subsec:preliminaries}, including horizon $H$. Appendix~\ref{appendix:adapters} describes environment restoration, early stopping, failure handling, and domain-specific guidance. Removing the hypergraph from the full agent also requires disabling any such guidance, beyond setting $\lambda=0$.

\section{Experiments}
\label{sec:exp}

We design our experiments around five research questions (RQs).
\textbf{RQ1 (Effectiveness):} does HyperMCTS improve end-to-end case accuracy over greedy reasoning and other search-based agents (Section~\ref{subsec:main})?
\textbf{RQ2 (Efficiency):} does it do so without inflating search cost in LLM calls and tokens (Section~\ref{subsec:efficiency})?
\textbf{RQ3 (Ablation):} how does removing the cross-trajectory hypergraph affect performance (Section~\ref{subsec:ablation})?
\textbf{RQ4 (Sensitivity):} how does performance vary with the hypergraph weight $\lambda$ (Section~\ref{subsec:sensitivity})?
\textbf{RQ5 (Generalizability):} does HyperMCTS generalize beyond planning to question answering over noisy evidence (Section~\ref{subsec:sealqa})?

\subsection{Experimental Setup}
\label{subsec:setup}

\textbf{Benchmarks.} DeepPlanning~\citep{zhang2026deepplanning} evaluates Travel Planning and Shopping Planning under verifiable constraints. Travel uses Commonsense Score (CS Score), Personalized Score (PS Score), and their mean, Composite Score (Comp Score). Shopping uses Match Score. Both domains report Case Accuracy, which requires all evaluation conditions to be satisfied; Average Accuracy is the mean Case Accuracy across the two domains. We additionally evaluate fact-seeking question answering over noisy evidence on SealQA~\citep{pham2025sealqa}, reporting answer accuracy and a partial-credit score (Mean) that assigns half credit to abstentions.

\textbf{Baselines.} We compare with ReAct~\citep{yao2023react}, unstructured TTS methods CoT~\citep{wei2022chain} and Reflexion~\citep{shinn2023reflexion}, and structured TTS methods ToT~\citep{yao2023tree}, RAP~\citep{hao2023reasoning}, LATS~\citep{zhou2024language}, and FLARE~\citep{wang2026flare}. RAP, LATS, and FLARE are MCTS variants.

\textbf{Models and implementation.} We use Qwen3.5-27B, Qwen3.6-27B, and Sonnet 4.6 in thinking mode, keeping the backbone model frozen and fixed within each comparison. Dataset details, baseline procedures, and search hyperparameters are provided in Appendix~\ref{appendix:setup}. Reward computation is described in Appendix~\ref{appendix:reward}.

\subsection{Main Results (RQ1)}
\label{subsec:main}

The DeepPlanning columns in Table~\ref{tab:main_results} report overall planning results, with full breakdowns in Appendix~\ref{appendix:breakdown}. Case Accuracy remains much lower than partial-credit scores. For example, Qwen3.6-27B ReAct attains $74.2\%$ Comp Score but only $10.4\%$ Case Accuracy on Travel. A successful plan must satisfy all constraints jointly, so one missed condition can invalidate otherwise correct decisions. All six TTS baselines improve Average Accuracy over ReAct on each model. Additional reasoning and exploration allow agents to revisit choices and detect missed constraints. Structured TTS generally yields larger gains in these comparisons. RAP achieves an Average Accuracy of $40.0\%$ on Qwen3.6-27B versus $36.7\%$ for Reflexion, while LATS reaches $39.2\%$ on Sonnet 4.6 versus $34.4\%$ for Reflexion. By explicitly retaining and evaluating alternatives, structured search can compare downstream consequences before selecting a plan.

\begin{table}[!t]
\centering
\caption{Results (\%) on DeepPlanning and SealQA. Avg Acc. averages Travel and Shopping Case Accuracy. \textbf{Bold} marks the best result per metric and model, including ties.}
\label{tab:main_results}
\small
\setlength{\tabcolsep}{3pt}
\begin{tabular*}{\linewidth}{@{\extracolsep{\fill}}*{10}{c}@{}}
\toprule
\multirow{3}{*}[-9.75pt]{Model} & \multirow{3}{*}[-9.75pt]{Agent} & \multicolumn{7}{c}{DeepPlanning} & SealQA \\
\cmidrule(lr){3-9}\cmidrule(lr){10-10}
 & & \multicolumn{4}{c}{Travel} & \multicolumn{2}{c}{Shopping} & \multirow{2}{*}[-7.5pt]{\tablehead{Avg\\Acc.}} & \multirow{2}{*}[-7.5pt]{Acc.} \\
\cmidrule(lr){3-6}\cmidrule(lr){7-8}
 & & \tablehead{CS\\Score} & \tablehead{PS\\Score} & \tablehead{Comp\\Score} & \tablehead{Case\\Acc.} & \tablehead{Match\\Score} & \tablehead{Case\\Acc.} & & \\
\midrule
\multirow{8}{*}[-6pt]{Qwen3.6-27B}
 & ReAct & 77.2 & 71.3 & 74.2 & 10.4 & 82.0  &  46.7  & 28.6 & 13.5 \\
\cmidrule[0.3pt](lr){2-10}
 & CoT & 76.2 & 67.1 & 71.7 & 11.7 & 83.6  &  51.7  & 31.7 & 14.4 \\
 & Reflexion & 82.0 & 75.4 & 78.7 & 21.7 & 84.8  &  51.7  & 36.7 & 14.4 \\
\cmidrule[0.3pt](lr){2-10}
 & ToT & 84.4 & 80.8 & 82.6 & 20.4 & 84.1  &  51.7  & 36.1 & 17.1 \\
 & RAP & 80.5 & 78.9 & 79.7 & 22.5 & 86.1  &  57.5  & 40.0 & 14.4 \\
 & LATS & 83.9 & 77.5 & 80.7 & 24.2 & 81.6  &  53.3  & 38.8 & 15.3 \\
 & FLARE & 79.9 & 77.0 & 78.4 & 20.0 & 83.2  &  53.3  & 36.7 & 18.9 \\
 & \textbf{HyperMCTS} & \textbf{86.9} & \textbf{83.8} & \textbf{85.3} & \textbf{33.8} & \textbf{87.0}  &  \textbf{60.8}  & \textbf{47.3} & \textbf{21.6} \\
\midrule
\multirow{8}{*}[-6pt]{Qwen3.5-27B}
 & ReAct & 68.2 & 68.3 & 68.3 & 7.1 & 82.8  &  49.2  & 28.2 & 12.6 \\
\cmidrule[0.3pt](lr){2-10}
 & CoT & 70.6 & 64.9 & 67.7 & 8.4 & 83.0  &  49.2  & 28.8 & 13.5 \\
 & Reflexion & 76.8 & 81.2 & 79.0 & 17.6 & 82.9  &  50.8  & 34.2 & 13.5 \\
\cmidrule[0.3pt](lr){2-10}
 & ToT & 76.4 & 76.5 & 76.5 & 15.4 & 83.2  &  50.8  & 33.1 & \textbf{17.1} \\
 & RAP & 75.0 & 78.7 & 76.8 & 15.0 & 82.2  &  50.8  & 32.9 & 13.5 \\
 & LATS & 76.9 & 78.6 & 77.7 & 17.5 & 83.1  &  50.0  & 33.8 & 14.4 \\
 & FLARE & 76.6 & 78.7 & 77.7 & 16.7 & 81.2  &  46.7  & 31.7 & 14.4 \\
 & \textbf{HyperMCTS} & \textbf{79.0} & \textbf{81.7} & \textbf{80.3} & \textbf{19.6} & \textbf{85.1}  &  \textbf{54.2}  & \textbf{36.9} & \textbf{17.1} \\
\midrule
\multirow{8}{*}[-6pt]{Sonnet 4.6}
 & ReAct & 75.9 & 64.2 & 70.0 & 13.4 & 76.6  &  41.7  & 27.6 & 14.4 \\
\cmidrule[0.3pt](lr){2-10}
 & CoT & 75.8 & 64.1 & 70.0 & 14.2 & 76.8  &  41.7  & 28.0 & 14.4 \\
 & Reflexion & 81.9 & 72.2 & 77.0 & 23.8 & 80.2  &  45.0  & 34.4 & 16.2 \\
\cmidrule[0.3pt](lr){2-10}
 & ToT & 82.1 & 70.0 & 76.0 & 21.3 & 78.0  &  43.3  & 32.3 & 22.5 \\
 & RAP & \textbf{86.4} & 74.2 & 80.3 & 29.6 & 81.9  &  48.3  & 39.0 & 17.1 \\
 & LATS & 84.9 & 78.4 & 81.6 & 29.2 & 83.1  &  49.2  & 39.2 & 19.8 \\
 & FLARE & 85.2 & \textbf{81.3} & \textbf{83.2} & \textbf{32.1} & 80.7  &  44.2  & 38.2 & 15.3 \\
 & \textbf{HyperMCTS} & 84.5 & \textbf{81.3} & 82.9 & 30.4 & \textbf{84.3}  &  \textbf{52.5}  & \textbf{41.5} & \textbf{26.1} \\
\bottomrule
\end{tabular*}
\end{table}

HyperMCTS attains an Average Accuracy of $47.3\%$, $36.9\%$, and $41.5\%$ on Qwen3.6-27B, Qwen3.5-27B, and Sonnet 4.6, respectively. The strongest baselines on these models are RAP ($40.0\%$), Reflexion ($34.2\%$), and LATS ($39.2\%$), giving HyperMCTS gains of $7.3$, $2.7$, and $2.3$ percentage points. On Travel, it leads all four metrics on both Qwen models, reaching $33.8\%$ Case Accuracy on Qwen3.6-27B versus $24.2\%$ for LATS. It also achieves the highest overall Travel PS Score across all three models, including a tie at $81.3\%$ on Sonnet 4.6. On Shopping, it leads both Match Score and Case Accuracy across all three models. With Qwen3.6-27B, it reaches $60.8\%$ Case Accuracy on Shopping, exceeding the $56.2\%$ reported for Claude Opus 4.6 (max) on the public leaderboard.\footnote{\url{https://qwenlm.github.io/Qwen-Agent/en/benchmarks/deepplanning/}} These joint gains reflect improvements in both satisfying individual requirements and producing fully correct plans.

\subsection{Search Efficiency (RQ2)}
\label{subsec:efficiency}

\begin{figure}[!t]
\centering
\includegraphics[width=0.88\linewidth]{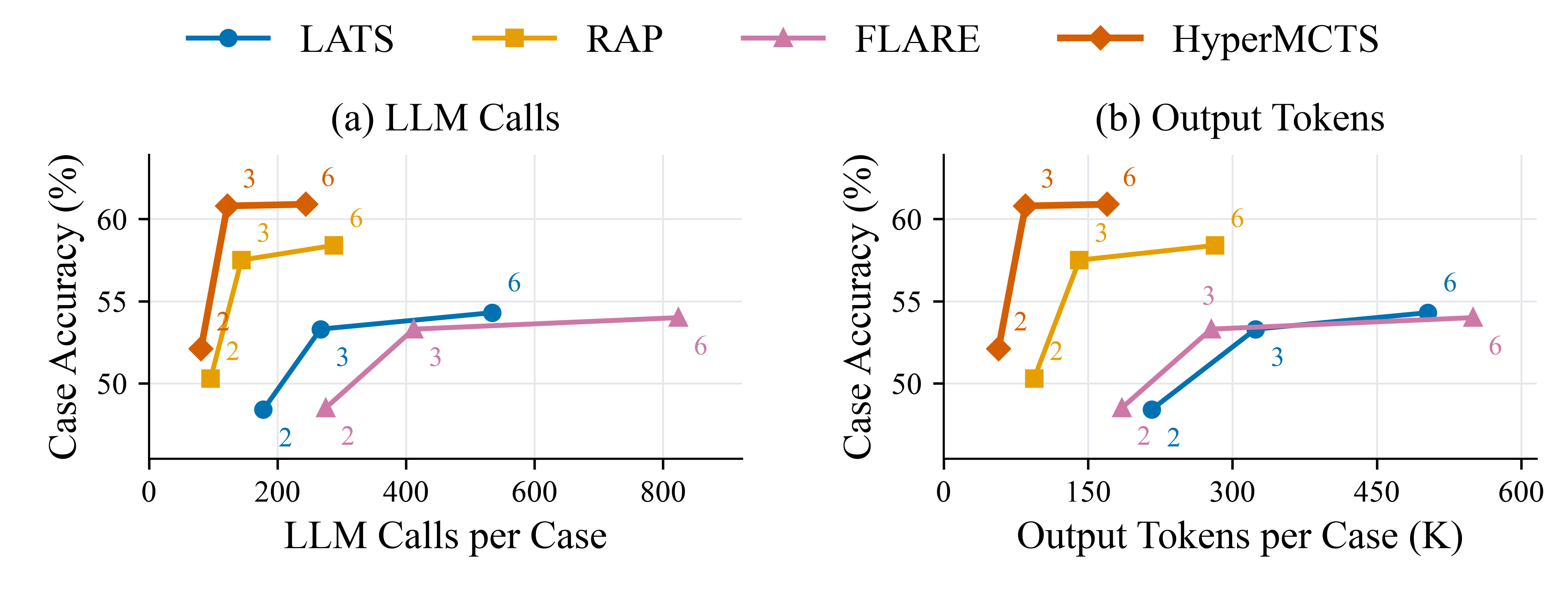}
\caption{Accuracy--cost trade-offs of MCTS-based methods on Shopping Planning ($120$ cases, Qwen3.6-27B, thinking mode). Costs are rounded per-case averages. Labels denote branching factors $2$, $3$, and $6$; lines connect configurations of the same method.}
\label{fig:efficiency}
\end{figure}

Figure~\ref{fig:efficiency} compares accuracy and per-case cost on Shopping Planning with Qwen3.6-27B in thinking mode. We evaluate RAP, LATS, FLARE, and HyperMCTS because these MCTS-based methods support the same branching-factor settings, $b\in\{2,3,6\}$. HyperMCTS achieves the highest Case Accuracy and uses the fewest LLM calls and output tokens at each setting. At $b=3$, it reaches $60.8\%$ accuracy with about $122$ calls and $85$K output tokens, while RAP reaches $57.5\%$ with $144$ calls and $141$K tokens. LATS and FLARE use $3.8\times$ and $3.3\times$ as many output tokens as HyperMCTS, respectively. FLARE requires the most LLM calls because it runs MCTS again from the updated state after each committed action to select the next action. Increasing $b$ from $3$ to $6$ yields only $0.1$--$1.0$ percentage points of accuracy gain across methods despite substantially higher costs. HyperMCTS at $b=3$ is more accurate and less costly than every baseline at $b=6$.

\subsection{Ablation Study (RQ3)}
\label{subsec:ablation}

\begin{figure}[!t]
\centering
\includegraphics[width=0.88\linewidth]{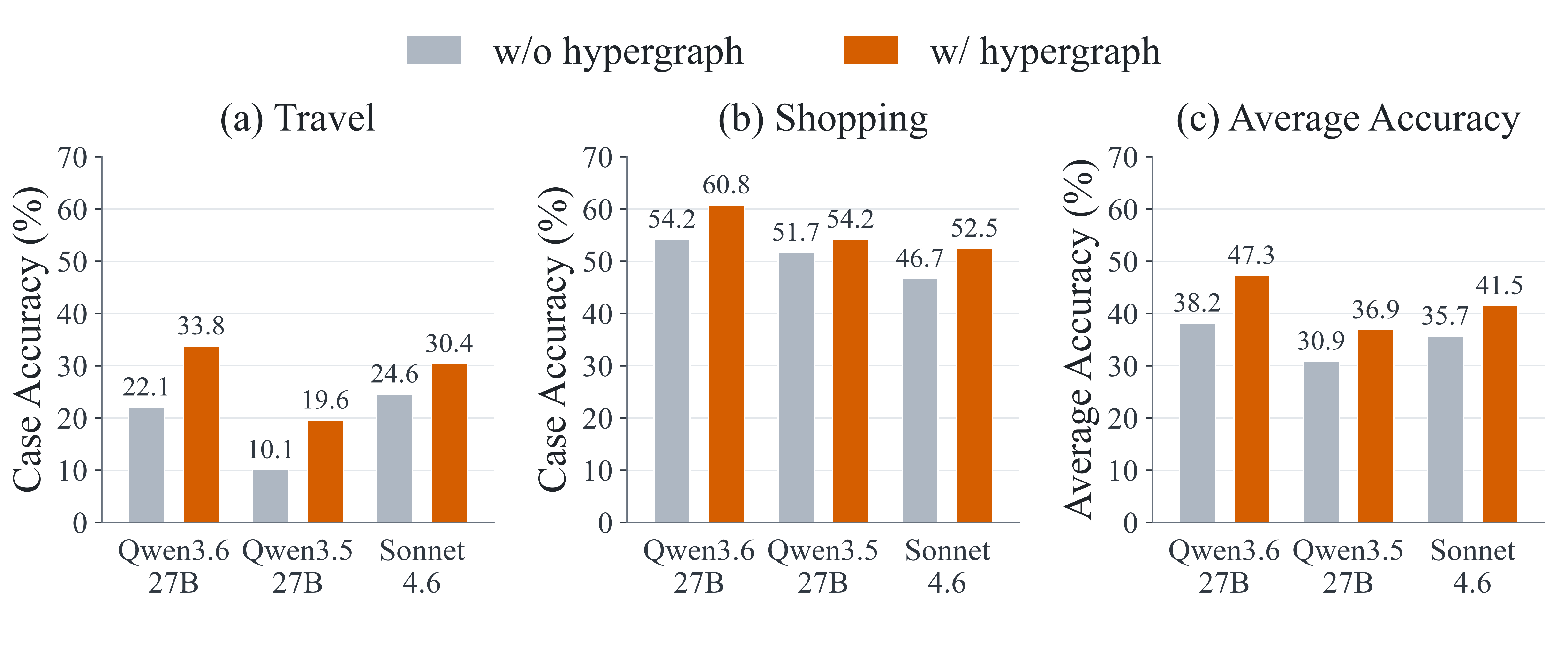}
\caption{Hypergraph ablation on DeepPlanning, comparing MCTS (w/o hypergraph) and HyperMCTS (w/ hypergraph).}
\label{fig:ablation}
\end{figure}

We compare HyperMCTS with the MCTS configuration without the hypergraph, using the same backbone models, rollout budget, and branching factor. Figure~\ref{fig:ablation} shows higher Case Accuracy for HyperMCTS in both domains on all three models. Average Accuracy is $47.3\%$ versus $38.2\%$ on Qwen3.6-27B, $36.9\%$ versus $30.9\%$ on Qwen3.5-27B, and $41.5\%$ versus $35.7\%$ on Sonnet 4.6. The two Qwen models benefit more on Travel, with gains of $11.7$ and $9.5$ percentage points, compared with $6.6$ and $2.5$ on Shopping. Sonnet gains $5.8$ points in each domain. This comparison evaluates the hypergraph component as a whole, rather than isolating the additive HyperUCT term from other uses of its statistics. More detailed results by language and difficulty are provided in Appendix~\ref{appendix:ablation}.

\subsection{Sensitivity Analysis (RQ4)}
\label{subsec:sensitivity}

\begin{figure}[!t]
\centering
\includegraphics[width=0.88\linewidth]{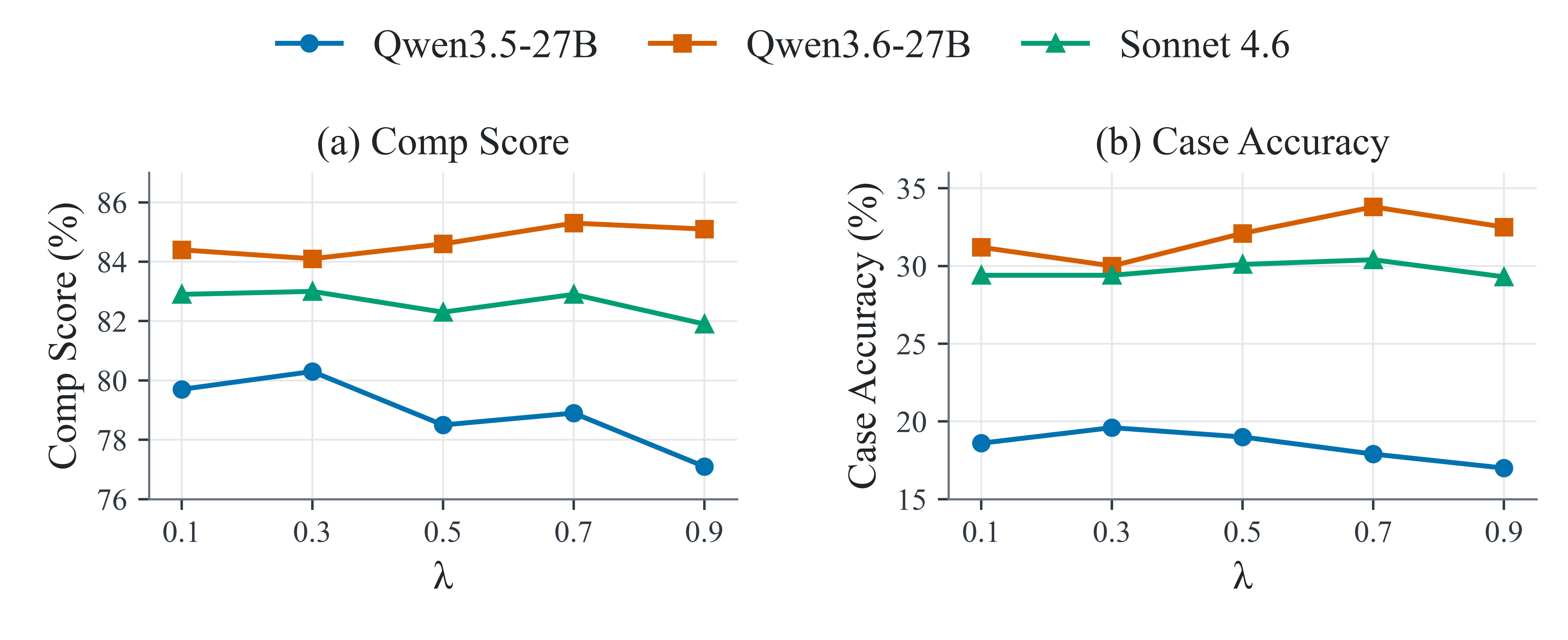}
\caption{Sensitivity to the hypergraph weight $\lambda$ on Travel Planning. Each curve shows one backbone model with $B=5$, $b=3$, and $c=1.414$.}
\label{fig:sensitivity}
\end{figure}

We vary the hypergraph weight $\lambda\in\{0.1,0.3,0.5,0.7,0.9\}$ on Travel while keeping $B$, $b$, and $c$ fixed. Figure~\ref{fig:sensitivity} shows that Qwen3.5-27B attains its highest Comp Score ($80.3\%$) and Case Accuracy ($19.6\%$) at $\lambda=0.3$. Qwen3.6-27B peaks at $\lambda=0.7$, reaching $85.3\%$ Comp Score and $33.8\%$ Case Accuracy. Sonnet also achieves its highest Case Accuracy ($30.4\%$) at $\lambda=0.7$, although its Comp Score peaks at $\lambda=0.3$. Stronger backbone models may produce more reliable search experience, allowing HyperMCTS to assign greater weight to cross-trajectory evidence in hypergraph statistics. Noisier experience from a weaker model may make a more moderate weight preferable. For the main Travel results, we use $\lambda=0.3$ for Qwen3.5-27B and $\lambda=0.7$ for Qwen3.6-27B and Sonnet 4.6. Full results are provided in Appendix~\ref{appendix:sensitivity}.

\FloatBarrier
\subsection{Generalizability (RQ5)}
\label{subsec:sealqa}

We evaluate on SealQA to test whether HyperMCTS generalizes beyond planning to question answering over noisy evidence. The SealQA column in Table~\ref{tab:main_results} reports accuracy, with full results in Appendix~\ref{appendix:sealqa} and the MCTS formulation in Appendix~\ref{appendix:sealqa_mcts}.

HyperMCTS achieves $21.6\%$ and $26.1\%$ accuracy on Qwen3.6-27B and Sonnet 4.6, exceeding the strongest baseline on each model by $2.7$ and $3.6$ percentage points, respectively. On Qwen3.5-27B, it ties ToT at $17.1\%$. These results extend the evaluation of HyperMCTS to question answering over noisy evidence.

\FloatBarrier

\section{Limitations}
\label{sec:limitations}

HyperMCTS assumes that the environment supports reversible actions, state restoration, or replay so that alternative action trajectories can be explored at test time. This is practical for constructing a shopping cart or revising a travel plan, where tentative choices can be changed before execution. Tasks with irreversible actions or persistent external effects require a simulator or another mechanism for evaluating alternatives without committing their effects.

HyperMCTS is intended for search trees with substantial overlap in decision groups across trajectories. Its hypergraph represents these groups as unordered sets, allowing their outcome statistics to be shared across different action prefixes. Limited overlap reduces the opportunities for such sharing. The set representation is also less suitable when outcomes depend strongly on action order or context omitted from the decision identities. Although the search tree preserves execution order, the hypergraph may pool incompatible feedback in these settings.

\section{Conclusion and Future Work}
\label{sec:conclusion}

HyperMCTS augments MCTS with a cross-trajectory hypergraph that organizes search outcomes by labelled decision groups. HyperUCT reads these statistics as an action-level prior alongside prefix-specific tree values and exploration. The resulting search retains ordered histories while making previously collected group evidence available across branches of the same task. On the evaluated DeepPlanning models, HyperMCTS achieves the highest Average Accuracy and improves over the no-hypergraph variant. The Shopping cost study shows higher accuracy with fewer calls and output tokens on Qwen3.6-27B, while the SealQA results extend the evaluation to question answering. Future work can combine HyperMCTS with world models to simulate action outcomes and support planning in agentic tasks with irreversible actions.

\bibliography{hypermcts}
\bibliographystyle{plainnat}

\appendix

\clearpage
\section{Additional Method Details}
\label{appendix:method}

\subsection{Notation Summary}
\label{appendix:notation}
Table~\ref{tab:notation} summarizes the notions used in the search tree and the cross-trajectory hypergraph.
\begin{table}[!t]
\centering
\caption{Core notation for the search tree and cross-trajectory hypergraph.}
\label{tab:notation}
\small
\begin{tabular}{lp{0.72\linewidth}}
\toprule
Symbol & Meaning \\
\midrule
$x$, $h_t$, $\tau$ & Query, observable interaction history, and complete or horizon-terminated trajectory \\
$H$, $B$, $b$ & Interaction horizon, search-iteration budget, and expansion branching factor \\
$\pi_\theta$, $\mathcal{A}_{\mathrm{dec}}(h)$ & Frozen proposal policy and task-defined search decisions \\
$V$, $R(\tau)$ & Terminal search verifier and trajectory reward \\
$\mathcal{G}$, $\mathcal{Y}$ & Query objectives and hyperedge labels \\
$\mathcal{U}$, $\phi$, $\mathcal{C}(\tau)$ & Canonical decision space, canonicalization map, and recorded decision set \\
$e=(S_e,y_e)$, $\Psi$ & Labelled decision-group record and observation extractor \\
$N_k(h)$ & Visit count of a history node \\
$N_k(h,a)$, $Q_k^{\mathrm{tree}}(h,a)$ & Tree-edge backup count and mean search feedback \\
$N_k^{\mathrm{hyper}}(e)$, $Q_k^{\mathrm{hyper}}(e)$ & Hyperedge observation count and mean feedback \\
$\overline{Q}_k^{\mathrm{hyper}}(v)$ & Action-level prior from incident hyperedges \\
$c$, $\lambda$ & UCT exploration coefficient and hypergraph-prior weight \\
\bottomrule
\end{tabular}
\end{table}

\subsection{Hypergraph Specification}
\label{appendix:hyperimpl}

This appendix separates the hypergraph's statistical interface from domain-specific search adapters. The general algorithm does not require the same low-level action vocabulary, recorded trace, or candidate-generation procedure in every task.

\paragraph{Decision identities.}
Canonicalization removes tree location while retaining task-relevant distinctions, such as product identifiers, tool arguments, dates, and semantic roles. Multiplicity and temporal roles must be encoded in the identity when relevant, since a set does not retain repeated occurrences. The recorded set $\mathcal{C}(\tau)$ may include only the decisions materialized in the tree; it need not include every tool call in the simulated suffix.

\paragraph{Observations and labels.}
The trajectory-level record associates the recorded decision set $\mathcal{C}(\tau)$ with $R(\tau)$ when the set contains at least two distinct identities. The objective-level instantiation associates each relevant decision $v$ with an objective $g$ through a singleton decision set, giving $e=(\{v\},g)$. These records also receive the terminal search score $R(\tau)$, rather than an independently measured objective reward. Although engineering implementations may call these records ``compatibility'' and ``satisfaction'' edges, those names do not establish compatibility in the current prefix or separate objective satisfaction. Multiple decision nodes are connected by the trajectory-level record.

\paragraph{Aggregation.}
Let $\mathcal{E}_{\mathrm{obj}}(v)$ and $\mathcal{E}_{\mathrm{traj}}(v)$ be the objective-level and trajectory-level records incident to $v$. Define $q_{\mathrm{obj}}(v)$ as the maximum empirical value over the former and $q_{\mathrm{traj}}(v)$ as the unweighted mean over the latter, when the corresponding set is nonempty. The aggregation is
\begin{equation}
\label{eq:implementation-agg}
\overline{Q}^{\mathrm{hyper}}(v)=
\begin{cases}
0.7q_{\mathrm{obj}}(v)+0.3q_{\mathrm{traj}}(v), & \text{both types present},\\
q_{\mathrm{obj}}(v), & \text{only objective records present},\\
q_{\mathrm{traj}}(v), & \text{only trajectory records present},\\
0, & \text{neither type present}.
\end{cases}
\end{equation}
Each incident edge contributes separately, even when another edge has the same label and value. This is a bounded heuristic. The maximum retains the strongest observed objective association; the trajectory term averages group means rather than weighting groups by their observation counts. Neither operation estimates the causal effect of selecting $v$.

\paragraph{Exact keys and approximate merging.}
The main formulation uses an exact key $(S_e,y_e)$. The implementation notes additionally describe a Jaccard-based union merge with threshold $0.8$. Jaccard similarity between two sets $S$ and $S'$ is $|S\cap S'|/|S\cup S'|$. Applying this operation to eligible records replaces their stored sets by $S\cup S'$. Such a record represents a cluster of overlapping observations; the union need not have appeared in a single rollout and its value is not the empirical return of that exact union. Its count and accumulated return must represent the contributing observations. A concrete merge policy must specify record-type and label eligibility, the treatment of objective metadata, and count migration. Online merge order and tie handling can affect the resulting clusters, so permutation invariance of an exact action-set key does not imply insertion-order invariance of approximate merging.

\subsection{Domain Interfaces and Search Integration}
\label{appendix:adapters}

\paragraph{Search-loop details.}
Alternative histories require environment replay, restorable snapshots, or a simulator. Terminal histories bypass expansion and simulation. If expansion yields no valid child, the adapter applies its task-specific failure rule. Candidate generation, child selection, ties, rollout depth, and stopping conditions are defined by the domain interface. When search stops before $B$ iterations, $\mathcal{D}_B$ denotes the trajectories evaluated before stopping; if none completes, the adapter returns its task-defined fallback. The final choice uses stored rewards without re-evaluation. Tree backup updates the represented path, while simulated suffix decisions may contribute hypergraph observations without becoming tree children.

\paragraph{Planning decisions.}
Shopping identifies product-addition decisions by product identifiers and associates them with parsed query requirements, such as a brand or a named item. Travel uses staged choices, including outbound transport, accommodation, daily plans, and return transport. Its recorded decision set can consist of the materialized tree path, with feedback obtained after completing the rest of the plan. This distinction is why Section~\ref{subsec:hypergraph} defines the recorded set separately from the full trajectory. Stage and temporal information are relevant to deciding which choices may share an identity.

\paragraph{SealQA as MCTS.}
\label{appendix:sealqa_mcts}
In the ReAct formulation of SealQA, a search state consists of the question and its accumulated action--observation history. The actions are \texttt{Search[query]}, which retrieves a passage from the question's document pool; \texttt{Lookup[keyword]}, which returns a matching sentence from the latest retrieved document; and \texttt{Finish[answer]}, which terminates the trajectory. Executing an action appends its observation to the history; alternative branches replay their action prefixes. Expansion proposes candidate continuations, and a bounded rollout completes the selected branch. Search feedback updates the tree path and hypergraph records over canonical action identities, with the question as the objective label. Action identities use the action type and arguments. An argument-only lookup identity can conflate decisions made under different retrieval contexts.

\paragraph{Search feedback and reporting.}
Appendix~\ref{appendix:reward} specifies the domain rewards and evaluation prompts. Evaluator and guidance calls contribute to computational cost even when they do not constitute another completed search iteration.

\clearpage
\subsection{Illustrative Search Example}
\label{appendix:case}

The shopping example in Figure~\ref{fig:overview} explains the update and readout operations. It is a constructed example, not an experimental case study. Panel (a) summarizes the decision sequences rather than specifying which tree nodes are materialized during simulation. The task is to select a monitor, keyboard and webcam within \$600. Monitors $A_1$ and $A_2$ cost \$400 and \$300; keyboard $B_1$ costs \$100; webcams $C_1$ and $C_2$ cost \$100 and \$200. The four rollouts $A_1B_1C_1$, $B_1A_1C_1$, $B_1A_1C_2$ and $B_1A_2C_2$ cost \$600, \$600, \$700 and \$600, respectively. These totals indicate budget feasibility only; $R_1$--$R_4$ denote symbolic search scores.

Under exact-key aggregation, the first two rollouts contribute two observations to $e_1=(\{A_1,B_1,C_1\},\mathrm{global})$, with $N^{\mathrm{hyper}}(e_1)=2$ and $Q^{\mathrm{hyper}}(e_1)=(R_1+R_2)/2$. The third and fourth rollouts update $e_2=(\{A_1,B_1,C_2\},\mathrm{global})$ and $e_3=(\{A_2,B_1,C_2\},\mathrm{global})$, each with count one and mean feedback $R_3$ and $R_4$, respectively. In the figure's hyperedge records, $N$ abbreviates $N^{\mathrm{hyper}}$.

At a later history $h$ after choosing $B_1$ then $A_1$, both $C_1$ and $C_2$ may be valid candidates. The prior for $C_1$ can read $e_1$, including feedback collected under prefix $A_1B_1$; the prior for $C_2$ can read both $e_2$ and $e_3$. Thus, the prior aggregates statistics from hyperedges incident to the candidate rather than filtering them by the current prefix. HyperUCT also uses their separate tree values and visit counts, so the hyperedge means alone do not determine which candidate is selected. This demonstrates a possible transfer of outcome evidence across prefixes. It does not establish how often that transfer occurs in the experiments, whether it changes a decision, or whether it prevents a repeated evaluation. The action identities, allowed construction orders, and illustrative outcomes must be checked against a concrete interface before using this example as an implementation trace.

\clearpage
\section{Implementation Details}
\label{appendix:implementation}

\subsection{Experimental and Implementation Details}
\label{appendix:setup}

\paragraph{Datasets and evaluation.}
DeepPlanning Travel contains $120$ tasks with Chinese and English versions, giving $240$ evaluated instances. Shopping contains $120$ multi-product cart tasks across three difficulty levels, with $50$, $50$, and $20$ cases at Levels 1, 2, and 3. Final outputs are scored by rule-based checkers; Travel additionally requires parsing the generated plan into a structured representation. Appendix~\ref{appendix:reward} specifies the feedback used during search and its relation to these reporting metrics. We use the $111$-question Seal-0 subset of SealQA; the accuracy and partial-credit formulas are given in Appendix~\ref{appendix:sealqa}. Planning scores and accuracies are reported as percentages, while SealQA Mean is on a $0$--$1$ scale. Table captions specify the boldface convention.

\paragraph{Baseline procedures.}
ReAct interleaves reasoning with tool actions and observations. CoT uses the same interaction procedure with an additional zero-shot instruction to reason step by step~\citep{kojima2022large}, while Reflexion uses verbal self-critique to revise subsequent attempts. These unstructured TTS methods do not maintain an explicit search tree. Structured TTS methods organize alternative reasoning or action trajectories, including ToT and the MCTS variants RAP, LATS, and FLARE. HyperMCTS belongs to the latter family and augments MCTS with a cross-trajectory hypergraph. All methods use the same frozen backbone model within each comparison. The open-weight Qwen3.5-27B and Qwen3.6-27B models and Sonnet 4.6 operate in thinking mode.

\paragraph{Search hyperparameters.}
Unless otherwise stated, HyperMCTS uses at most $B=5$ search iterations and branching factor $b=3$ on DeepPlanning. We use a UCT exploration coefficient $c=1.414$ as in~\citet{park2025ensembling} and~\citet{chen2026mist}. The efficiency study varies $b\in\{2,3,6\}$, as described in Section~\ref{subsec:efficiency}.

\paragraph{Reported weights.}
The Shopping configuration uses $\lambda=0.5$. Travel uses $\lambda=0.3$ for Qwen3.5-27B and $\lambda=0.7$ for Qwen3.6-27B and Sonnet 4.6 (Section~\ref{subsec:sensitivity}). SealQA uses $\lambda=0.2$ for Qwen3.6-27B and Sonnet 4.6 and $\lambda=0.5$ for Qwen3.5-27B and Opus 4.8, as detailed in Appendix~\ref{appendix:sealqa}.

\clearpage
\subsection{Reward Computation and Evaluation Prompts}
\label{appendix:reward}

We describe reward computation for DeepPlanning and SealQA. The hypergraph prior guides selection but is not added to the reward. Terminal rewards update both tree and hypergraph statistics.

\paragraph{Travel Planning.}
We convert the final itinerary to JSON using DeepPlanning's language-specific conversion prompts~\citep{zhang2026deepplanning}, with the parser configured under the \texttt{claude-haiku} alias. The parser formats the plan; programmatic checkers evaluate it against task constraints and the local travel database. For parsed plan $p_\tau$,
\begin{equation}
\label{eq:travel-reward}
R_{\mathrm{travel}}(\tau)=\tfrac{1}{2}\bigl[\mathrm{CS}(p_\tau)+\mathrm{PS}(p_\tau)\bigr],
\qquad \mathrm{CS}(p_\tau)=\tfrac{1}{8}\sum_{d=1}^{8}c_d(p_\tau).
\end{equation}
Each commonsense dimension $c_d$ is one only if its nonempty set of evaluated checks all pass, and zero otherwise. PS applies the same all-pass rule to the personalized constraints. Undefined checks are excluded. Plans shorter than 50 characters, failed conversion, a missing task database, or checker errors receive zero. If the database argument or sample identifier is absent, a separate fallback returns $0.3$ when the parsed object contains \texttt{daily\_plans}, and $0.1$ otherwise.

\paragraph{Shopping Planning.}
The judge uses the agent's model configuration and receives the request, cached product attributes, and applied coupons. Serialized product details are truncated to 3,000 characters; the full interaction transcript is not provided. For a valid JSON response,
\begin{equation}
\label{eq:shopping-reward}
R_{\mathrm{shopping}}(\tau)=\operatorname{clip}_{[0,1]}(s_\tau),
\end{equation}
where $s_\tau$ is the returned \texttt{score}, used directly rather than recomputed from the returned counts. Empty carts receive zero. Failed JSON extraction triggers a textual-fraction fallback, returning the ratio (zero for a zero denominator) without clipping; this fallback does not enforce the normalized-reward assumption in Section~\ref{subsec:preliminaries}. Unparseable responses or judge-call errors return $0.5$. Final benchmark scores instead use reference-product and coupon matching.

\paragraph{SealQA.}
A trajectory-value evaluator uses the actor's model configuration to score the question and its sequence of thoughts, actions, and observations. Its LATS-style prompt~\citep{zhou2024language} requests an integer $s_\tau\in\{1,\ldots,10\}$, giving
\begin{equation}
\label{eq:sealqa-reward}
R_{\mathrm{SealQA}}(\tau)=s_\tau/10.
\end{equation}
The evaluator also scores intermediate candidates; the terminal \texttt{value\_score} is used for backup and hypergraph updates. Empty trajectories or failed calls receive zero. If numeric parsing fails, responses containing ``incorrect'' or ``wrong'' receive $0.2$; otherwise, ``correct'' or ``reasonable'' yields $0.7$, and other responses yield zero. A separate reference-answer grader maps \texttt{CORRECT}, \texttt{NOT\_ATTEMPTED}, and \texttt{INCORRECT} to $1$, $0.5$, and $0$ for reporting. For SealQA, the grader's score and label remain in the terminal observation supplied to the value evaluator, including after forced \texttt{Finish}. Thus, terminal value evaluation receives reference-derived feedback even though the value prompt has no reference-answer field.

All three configurations retain the candidate with the highest observed terminal reward. Travel and Shopping permit early stopping at $0.95$; SealQA uses its configured \texttt{value\_stop}. The complete search-evaluation prompts follow, with braces denoting runtime substitutions. Travel reuses DeepPlanning's conversion prompts and has no additional LLM scoring prompt.

\noindent\fcolorbox{black!30}{black!2}{%
\begin{minipage}{\dimexpr\linewidth-2\fboxsep-2\fboxrule\relax}
\textbf{Shopping terminal-judge prompt}\par\medskip
\label{appendix:shopping_reward_prompt}
\begingroup
\fontsize{9}{10.5}\fontencoding{T1}\selectfont\rmfamily\raggedright
\hyphenpenalty=10000\exhyphenpenalty=10000
\setlength{\parindent}{0pt}\setlength{\parskip}{0pt}
\setlength{\leftskip}{6pt}\setlength{\rightskip}{6pt}
You are a strict evaluator checking whether a shopping cart perfectly satisfies a user\textquotesingle{}s requirements.\par
\smallskip
\#\# Evaluation Rules\par
- Each requirement must be satisfied by EXACTLY ONE product in the cart\par
- ALL constraints in a requirement must be met (brand, name keywords, color, size, rating thresholds, sales volume, stock, delivery time, etc.)\par
- If a requirement says \textquotedbl{}more than 200 four-star reviews\textquotedbl{}, check the actual number \textemdash{} 199 does NOT satisfy it\par
- If a requirement says \textquotedbl{}\textquotesingle{}Omni-Wick\textquotesingle{} in the name\textquotedbl{}, the product name MUST contain that exact phrase\par
- If a requirement specifies a brand, ONLY that brand satisfies it \textemdash{} no substitutes\par
- A product that meets MOST but not ALL constraints of a requirement counts as NOT satisfied\par
- Be strict: when in doubt, mark as NOT satisfied\par
\smallskip
\#\# Few-Shot Examples\par
\smallskip
Example 1:\par
Requirement: \textquotedbl{}a pink \textquotesingle{}Active Performance Tank Top\textquotesingle{} from Puma with average score \textgreater{} 4.5\textquotedbl{}\par
Product in cart: name=\textquotedbl{}Puma Active Performance Tank Top\textquotedbl{}, brand=\textquotedbl{}Puma\textquotedbl{}, color=\textquotedbl{}Pink\textquotedbl{}, avg\_score=4.7\par
\ensuremath{\rightarrow} SATISFIED (brand \ding{51}, name contains phrase \ding{51}, color \ding{51}, score 4.7\textgreater{}4.5 \ding{51})\par
\smallskip
Example 2:\par
Requirement: \textquotedbl{}a pink \textquotesingle{}Active Performance Tank Top\textquotesingle{} from Puma with average score \textgreater{} 4.5\textquotedbl{}\par
Product in cart: name=\textquotedbl{}Puma Classic Running Vest\textquotedbl{}, brand=\textquotedbl{}Puma\textquotedbl{}, color=\textquotedbl{}Pink\textquotedbl{}, avg\_score=4.8\par
\ensuremath{\rightarrow} NOT SATISFIED (name does NOT contain \textquotesingle{}Active Performance Tank Top\textquotesingle{} \ding{55})\par
\smallskip
Example 3:\par
Requirement: \textquotedbl{}item from Salomon with stock quantity over 250 and more than 400 five-star ratings\textquotedbl{}\par
Product in cart: name=\textquotedbl{}Salomon Trail Shoes\textquotedbl{}, brand=\textquotedbl{}Salomon\textquotedbl{}, stock=300, five\_star=380\par
\ensuremath{\rightarrow} NOT SATISFIED (five\_star 380 \textless{} 400 \ding{55}, even though stock 300\textgreater{}250 \ding{51})\par
\smallskip
Example 4:\par
Requirement: \textquotedbl{}monthly sales volume more than 550, transport time less than 2 days\textquotedbl{}\par
Product in cart: monthly\_sales=600, transport\_time=3 days\par
\ensuremath{\rightarrow} NOT SATISFIED (transport\_time 3 \textgreater{} 2 \ding{55})\par
\smallskip
\#\# User\textquotesingle{}s Shopping Request\par
\{self.query\}\par
\smallskip
\#\# Current Cart Contents\par
\{json.dumps(cart\_details, indent=2, ensure\_ascii=False)[:3000]\}\par
\smallskip
\#\# Coupons Applied\par
\{json.dumps(used\_coupons, ensure\_ascii=False) if used\_coupons else \textquotesingle{}None\textquotesingle{}\}\par
\smallskip
\#\# Task\par
For each requirement, check if there is a product in the cart that satisfies ALL its constraints.\par
Be strict \textemdash{} partial matches do not count.\par
\smallskip
\#\# Output\par
Return ONLY a JSON object:\par
\{\textquotedbl{}satisfied\_count\textquotedbl{}: \textless{}int\textgreater{}, \textquotedbl{}total\_requirements\textquotedbl{}: \textless{}int\textgreater{}, \textquotedbl{}score\textquotedbl{}: \textless{}float 0-1\textgreater{}, \textquotedbl{}reasoning\textquotedbl{}: \textquotedbl{}\textless{}brief explanation per requirement\textgreater{}\textquotedbl{}\}\par
\endgroup
\end{minipage}%
}\par\medskip

\noindent\fcolorbox{black!30}{black!2}{%
\begin{minipage}{\dimexpr\linewidth-2\fboxsep-2\fboxrule\relax}
\textbf{SealQA trajectory-value prompt}\par\medskip
\label{appendix:sealqa_reward_prompt}
\begingroup
\fontsize{9}{10.5}\fontencoding{T1}\selectfont\rmfamily\raggedright
\hyphenpenalty=10000\exhyphenpenalty=10000
\setlength{\parindent}{0pt}\setlength{\parskip}{0pt}
\setlength{\leftskip}{6pt}\setlength{\rightskip}{6pt}
Analyze the trajectory of a solution to a fact-seeking question answering task. The trajectory consists of:\par
\hspace*{1.0em}- Thought N: reasoning about the current situation\par
\hspace*{1.0em}- Action N: one of Search[entity], Lookup[keyword], or Finish[answer]\par
\hspace*{1.0em}- Observation N: the result of executing the action\par
\smallskip
Given the question and the trajectory, evaluate its correctness. Pay particular attention to whether:\par
\hspace*{1.0em}- The retrieved evidence actually supports the final answer\par
\hspace*{1.0em}- The final answer is consistent with the observations and not a hallucination\par
\hspace*{1.0em}- The trajectory navigated noisy/conflicting search results sensibly\par
\hspace*{1.0em}- For false-premise questions, whether the trajectory recognizes the false premise\par
\smallskip
Provide brief reasoning, then on the LAST line conclude exactly:\par
\textquotedbl{}Thus the correctness score is s\textquotedbl{}\par
where s is an integer from 1 to 10 (10 = clearly correct, 1 = clearly wrong).\par
\smallskip
Question: \{question\}\par
\smallskip
Trajectory:\par
\{trajectory\}\par
\smallskip
Reasoning:\par
\endgroup
\end{minipage}%
}\par\medskip

\clearpage
\section{Additional Experimental Results}
\label{appendix:breakdown}

This appendix provides the complete per-language Travel Planning and per-difficulty Shopping Planning breakdowns underlying the DeepPlanning results in Table~\ref{tab:main_results}, including Qwen3.6-27B, Qwen3.5-27B, and Sonnet 4.6. It also reports detailed ablation and sensitivity results, along with the full SealQA metrics and additional results for Opus 4.8.

\subsection{Travel Planning}

Table~\ref{tab:travel_breakdown} reports the detailed Travel Planning results for all three models. On Sonnet 4.6, HyperMCTS achieves the highest EN PS Score ($88.3\%$) and Comp Score ($91.8\%$); its overall PS Score is $81.3\%$ (tied best), and its overall Comp Score is $82.9\%$.

Sonnet 4.6 performs substantially better on EN than ZH in Travel, with HyperMCTS attaining $53.3\%$ versus $7.5\%$ Case Accuracy, as shown in Table~\ref{tab:travel_breakdown}. This contrast with Qwen's stronger ZH results may partly reflect differences in Chinese--English capabilities acquired during pretraining, which can influence how reliably each model interprets and satisfies planning constraints.

\begin{table}[!t]
\centering
\caption{Travel Planning, per-language breakdown (\%) with all models in thinking mode. CS Score (Commonsense Score), PS Score (Personalized Score), Comp Score (Composite Score), and Case Acc. (Case Accuracy) follow the definitions in the main text. For each model and language group (including Overall), the best value of every metric is in \textbf{bold}, including ties.}
\label{tab:travel_breakdown}
\begingroup
\fontsize{9}{10.5}\selectfont
\setlength{\tabcolsep}{1pt}
\begin{tabular*}{\linewidth}{@{\extracolsep{\fill}}*{14}{c}@{}}
\toprule
\multirow{2}{*}[-7.5pt]{Model} & \multirow{2}{*}[-7.5pt]{Agent} & \multicolumn{4}{c}{ZH} & \multicolumn{4}{c}{EN} & \multicolumn{4}{c}{Overall} \\
\cmidrule(lr){3-6} \cmidrule(lr){7-10} \cmidrule(lr){11-14}
 & & \tablehead{CS\\Score} & \tablehead{PS\\Score} & \tablehead{Comp\\Score} & \tablehead{Case\\Acc.} & \tablehead{CS\\Score} & \tablehead{PS\\Score} & \tablehead{Comp\\Score} & \tablehead{Case\\Acc.} & \tablehead{CS\\Score} & \tablehead{PS\\Score} & \tablehead{Comp\\Score} & \tablehead{Case\\Acc.} \\
\midrule
\multirow{8}{*}[-6pt]{Qwen3.6-27B}
 & ReAct & 74.7 & 76.7 & 75.7 & 7.5 & 79.6 & 65.8 & 72.7 & 13.3 & 77.2 & 71.3 & 74.2 & 10.4 \\
\cmidrule[0.3pt](lr){2-14}
 & CoT & 73.3 & 72.5 & 72.9 & 6.7 & 79.1 & 61.7 & 70.4 & 16.7 & 76.2 & 67.1 & 71.7 & 11.7 \\
 & Reflexion & 79.5 & 85.8 & 82.7 & 15.8 & 84.5 & 65.0 & 74.8 & 27.5 & 82.0 & 75.4 & 78.7 & 21.7 \\
\cmidrule[0.3pt](lr){2-14}
 & ToT & 82.4 & 86.6 & 84.5 & 15.8 & 86.4 & \textbf{75.0} & 80.7 & 25.0 & 84.4 & 80.8 & 82.6 & 20.4 \\
 & RAP & 79.7 & 82.8 & 81.3 & 22.5 & 81.2 & \textbf{75.0} & 78.1 & 22.5 & 80.5 & 78.9 & 79.7 & 22.5 \\
 & LATS & 81.7 & 85.8 & 83.8 & 22.5 & 86.0 & 69.2 & 77.6 & 25.8 & 83.9 & 77.5 & 80.7 & 24.2 \\
 & FLARE & 78.9 & 83.0 & 81.0 & 18.3 & 80.8 & 71.0 & 75.9 & 21.7 & 79.9 & 77.0 & 78.4 & 20.0 \\
 & HyperMCTS & \textbf{85.5} & \textbf{93.3} & \textbf{89.4} & \textbf{29.2} & \textbf{88.3} & 74.2 & \textbf{81.3} & \textbf{38.3} & \textbf{86.9} & \textbf{83.8} & \textbf{85.3} & \textbf{33.8} \\
\midrule
\multirow{8}{*}[-6pt]{Qwen3.5-27B}
 & ReAct & 67.6 & 68.3 & 68.0 & 6.7 & 68.8 & 68.3 & 68.6 & 7.5 & 68.2 & 68.3 & 68.3 & 7.1 \\
\cmidrule[0.3pt](lr){2-14}
 & CoT & 72.0 & 68.9 & 70.5 & 8.4 & 69.2 & 60.8 & 65.0 & 8.3 & 70.6 & 64.9 & 67.7 & 8.4 \\
 & Reflexion & 76.7 & 84.9 & 80.8 & 17.7 & 76.8 & \textbf{77.5} & 77.2 & 17.5 & 76.8 & 81.2 & 79.0 & 17.6 \\
\cmidrule[0.3pt](lr){2-14}
 & ToT & 75.6 & 78.8 & 77.2 & 13.3 & 77.2 & 74.2 & 75.7 & 17.5 & 76.4 & 76.5 & 76.5 & 15.4 \\
 & RAP & 74.9 & 81.5 & 78.2 & 14.2 & 75.1 & 75.8 & 75.5 & 15.8 & 75.0 & 78.7 & 76.8 & 15.0 \\
 & LATS & 75.4 & 81.4 & 78.4 & 16.7 & 78.3 & 75.8 & 77.1 & 18.3 & 76.9 & 78.6 & 77.7 & 17.5 \\
 & FLARE & 76.2 & 81.3 & 78.8 & 16.7 & 77.0 & 76.1 & 76.6 & 16.7 & 76.6 & 78.7 & 77.7 & 16.7 \\
 & HyperMCTS & \textbf{79.0} & \textbf{86.4} & \textbf{82.7} & \textbf{20.3} & \textbf{79.0} & 76.9 & \textbf{78.0} & \textbf{18.8} & \textbf{79.0} & \textbf{81.7} & \textbf{80.3} & \textbf{19.6} \\
\midrule
\multirow{8}{*}[-6pt]{Sonnet 4.6}
 & ReAct & 65.7 & 55.1 & 60.4 & 1.7 & 86.0 & 73.3 & 79.7 & 25.0 & 75.9 & 64.2 & 70.0 & 13.4 \\
\cmidrule[0.3pt](lr){2-14}
 & CoT & 66.1 & 55.0 & 60.6 & 2.5 & 85.5 & 73.2 & 79.4 & 25.8 & 75.8 & 64.1 & 70.0 & 14.2 \\
 & Reflexion & 74.5 & 69.9 & 72.2 & \textbf{7.5} & 89.3 & 74.4 & 81.9 & 40.0 & 81.9 & 72.2 & 77.0 & 23.8 \\
\cmidrule[0.3pt](lr){2-14}
 & ToT & 73.1 & 67.5 & 70.3 & 4.2 & 91.0 & 72.5 & 81.8 & 38.3 & 82.1 & 70.0 & 76.0 & 21.3 \\
 & RAP & \textbf{77.1} & 71.7 & 74.4 & 5.8 & \textbf{95.6} & 76.7 & 86.2 & 53.3 & \textbf{86.4} & 74.2 & 80.3 & 29.6 \\
 & LATS & 75.6 & \textbf{76.7} & \textbf{76.2} & \textbf{7.5} & 94.1 & 80.0 & 87.1 & 50.8 & 84.9 & 78.4 & 81.6 & 29.2 \\
 & FLARE & 74.9 & \textbf{76.7} & 75.8 & \textbf{7.5} & 95.5 & 85.8 & 90.7 & \textbf{56.7} & 85.2 & \textbf{81.3} & \textbf{83.2} & \textbf{32.1} \\
 & HyperMCTS & 73.7 & 74.2 & 74.0 & \textbf{7.5} & 95.2 & \textbf{88.3} & \textbf{91.8} & 53.3 & 84.5 & \textbf{81.3} & 82.9 & 30.4 \\
\bottomrule
\end{tabular*}
\endgroup
\end{table}

\clearpage
\subsection{Shopping Planning}

Table~\ref{tab:shopping_breakdown} reports the detailed Shopping Planning results for all three models.

\begin{table}[!t]
\centering
\caption{Shopping Planning, per-difficulty breakdown (\%) with all models in thinking mode. Case Acc. denotes Case Accuracy. For each model and difficulty group (including Overall), the best value of every metric is in \textbf{bold}, including ties.}
\label{tab:shopping_breakdown}
\setlength{\tabcolsep}{5pt}
\begin{tabular}{*{8}{c}}
\toprule
\multirow{2}{*}[-8pt]{Level} & \multirow{2}{*}[-8pt]{Agent} & \multicolumn{2}{c}{\textbf{Qwen3.6-27B}} & \multicolumn{2}{c}{\textbf{Qwen3.5-27B}} & \multicolumn{2}{c}{\textbf{Sonnet 4.6}} \\
\cmidrule(lr){3-4} \cmidrule(lr){5-6} \cmidrule(lr){7-8}
 & & \tablehead{Match\\Score} & \tablehead{Case\\Acc.} & \tablehead{Match\\Score} & \tablehead{Case\\Acc.} & \tablehead{Match\\Score} & \tablehead{Case\\Acc.} \\
\midrule
\multirow{8}{*}[-6pt]{L1 ($n{=}50$)}
 & ReAct      & 84.7 & 46.0 & 83.3 & 46.0 & 75.3 & 36.0 \\
\cmidrule[0.3pt](lr){2-8}
 & CoT       & 84.6 & 50.0 & \textbf{86.1} & \textbf{52.0} & 75.8 & 38.0 \\
 & Reflexion & 86.5 & 54.0 & 84.2 & \textbf{52.0} & 84.2 & 46.0 \\
\cmidrule[0.3pt](lr){2-8}
 & ToT       & 86.1 & 50.0 & 84.2 & 48.0 & 76.0 & 38.0 \\
 & RAP       & 88.4 & 58.0 & 82.8 & 46.0 & 81.4 & 44.0 \\
 & LATS      & 84.2 & 54.0 & \textbf{86.1} & \textbf{52.0} & 84.7 & 46.0 \\
 & FLARE     & 87.0 & 58.0 & 83.2 & 46.0 & 80.1 & 42.0 \\
 & HyperMCTS & \textbf{88.8} & \textbf{62.0} & 85.1 & 50.0 & \textbf{85.6} & \textbf{50.0} \\
\midrule
\multirow{8}{*}[-6pt]{L2 ($n{=}50$)}
 & ReAct      & 80.0 & 48.0 & 80.9 & 52.0 & 77.0 & 48.0 \\
\cmidrule[0.3pt](lr){2-8}
 & CoT       & \textbf{83.9} & \textbf{56.0} & 78.3 & 42.0 & 77.0 & 46.0 \\
 & Reflexion & 82.2 & 46.0 & 81.7 & 50.0 & 75.2 & 42.0 \\
\cmidrule[0.3pt](lr){2-8}
 & ToT       & 82.6 & 54.0 & 81.3 & 52.0 & 79.6 & 48.0 \\
 & RAP       & 83.5 & 54.0 & 77.8 & 46.0 & 80.9 & 54.0 \\
 & LATS      & 79.1 & 52.0 & 79.6 & 48.0 & 81.3 & 52.0 \\
 & FLARE     & 80.0 & 46.0 & 78.0 & 46.0 & 79.9 & 48.0 \\
 & HyperMCTS & \textbf{83.9} & \textbf{56.0} & \textbf{84.3} & \textbf{54.0} & \textbf{85.2} & \textbf{58.0} \\
\midrule
\multirow{8}{*}[-6pt]{L3 ($n{=}20$)}
 & ReAct      & 80.4 & 45.0 & 86.1 & 50.0 & 78.6 & 40.0 \\
\cmidrule[0.3pt](lr){2-8}
 & CoT       & 80.2 & 45.0 & 87.1 & 60.0 & 78.6 & 40.0 \\
 & Reflexion & 87.1 & 60.0 & 82.9 & 50.0 & 82.9 & \textbf{50.0} \\
\cmidrule[0.3pt](lr){2-8}
 & ToT       & 82.8 & 50.0 & 85.7 & 55.0 & 79.1 & 45.0 \\
 & RAP       & 87.1 & 65.0 & \textbf{91.4} & \textbf{75.0} & \textbf{85.7} & 45.0 \\
 & LATS      & 81.2 & 55.0 & 84.2 & 50.0 & 83.5 & \textbf{50.0} \\
 & FLARE     & 81.4 & 60.0 & 84.3 & 50.0 & 84.0 & 40.0 \\
 & HyperMCTS & \textbf{90.0} & \textbf{70.0} & 87.1 & 65.0 & 78.6 & 45.0 \\
\midrule
\multirow{8}{*}[-6pt]{Overall ($n{=}120$)}
 & ReAct      & 82.0 & 46.7 & 82.8 & 49.2 & 76.6 & 41.7 \\
\cmidrule[0.3pt](lr){2-8}
 & CoT       & 83.6 & 51.7 & 83.0 & 49.2 & 76.8 & 41.7 \\
 & Reflexion & 84.8 & 51.7 & 82.9 & 50.8 & 80.2 & 45.0 \\
\cmidrule[0.3pt](lr){2-8}
 & ToT       & 84.1 & 51.7 & 83.2 & 50.8 & 78.0 & 43.3 \\
 & RAP       & 86.1 & 57.5 & 82.2 & 50.8 & 81.9 & 48.3 \\
 & LATS      & 81.6 & 53.3 & 83.1 & 50.0 & 83.1 & 49.2 \\
 & FLARE     & 83.2 & 53.3 & 81.2 & 46.7 & 80.7 & 44.2 \\
 & HyperMCTS & \textbf{87.0} & \textbf{60.8} & \textbf{85.1} & \textbf{54.2} & \textbf{84.3} & \textbf{52.5} \\
\bottomrule
\end{tabular}
\end{table}

\clearpage
\subsection{Hypergraph Ablation Breakdowns}
\label{appendix:ablation}

\begin{figure}[!t]
\centering
\includegraphics[width=0.98\linewidth]{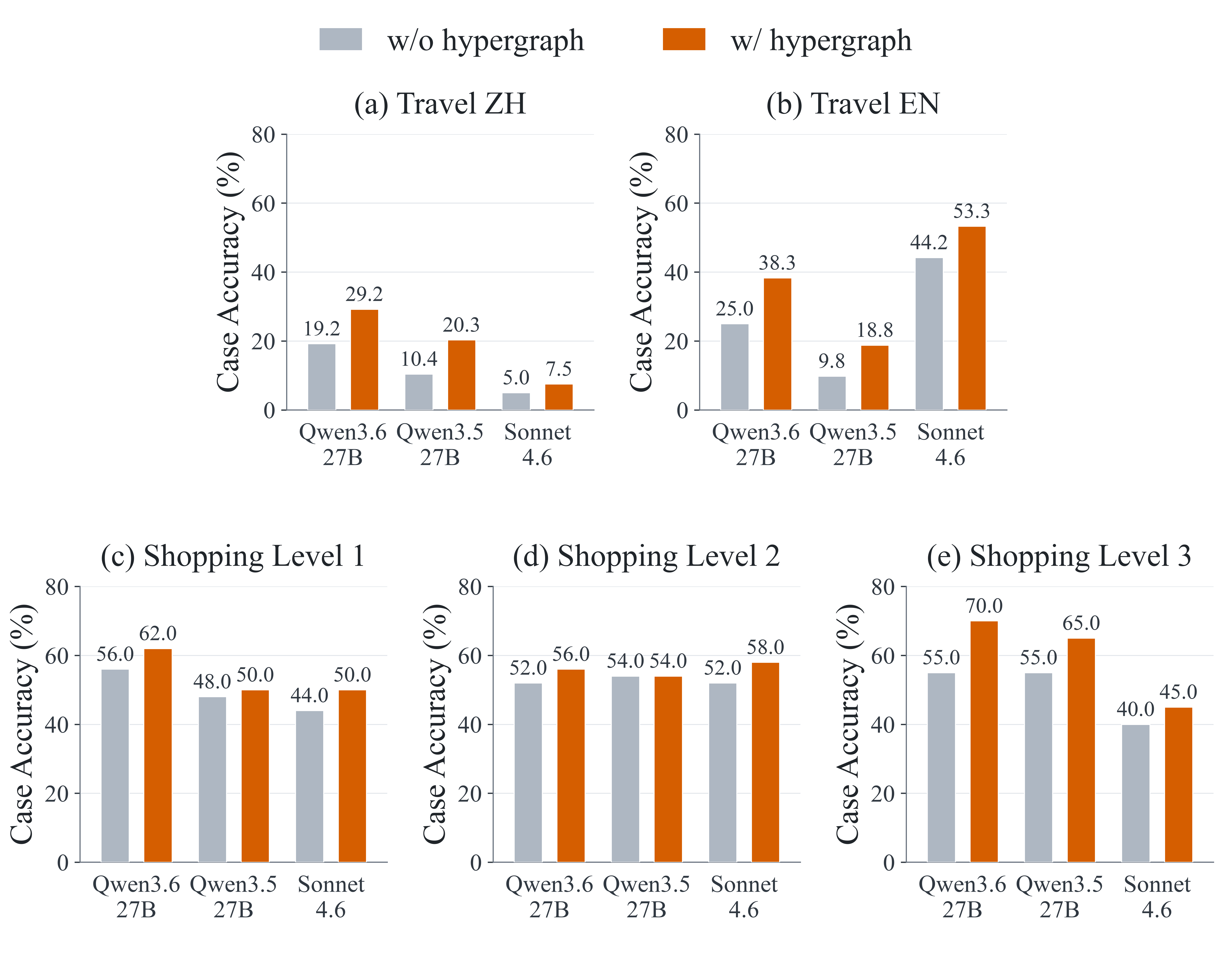}
\caption{Case Accuracy (\%) by Travel language and Shopping difficulty for MCTS (w/o hypergraph) and HyperMCTS (w/ hypergraph).}
\label{fig:ablation_breakdown}
\end{figure}

Figure~\ref{fig:ablation_breakdown} reports Case Accuracy by language and difficulty for the ablation in Section~\ref{subsec:ablation}. HyperMCTS improves over MCTS on both ZH and EN for all three models. On Shopping, the largest gains for the two Qwen models occur at Level 3, which includes coupon optimization, reaching $15.0$ and $10.0$ percentage points. Sonnet's gains are similar across levels ($6.0$, $6.0$, and $5.0$ points). Qwen3.5-27B ties MCTS at $54.0\%$ on Level 2. These results suggest that the benefit depends on both the model and the task; it does not increase uniformly with difficulty. Level 3 contains only $20$ cases, compared with $50$ each for Levels 1 and 2.

\clearpage
\subsection{Hypergraph Weight Sensitivity}
\label{appendix:sensitivity}

Table~\ref{tab:lambda_sensitivity} reports all four overall Travel metrics for the sweep in Section~\ref{subsec:sensitivity}. We hold $B=5$, $b=3$, and $c=1.414$ fixed across the five values of $\lambda$.

\begin{table}[!t]
\centering
\caption{Travel sensitivity to $\lambda$ (\%). \textbf{Bold} marks the best value of each metric within each model, including ties.}
\label{tab:lambda_sensitivity}
\setlength{\tabcolsep}{8pt}
\begin{tabular}{*{6}{c}}
\toprule
Model & $\lambda$ & \tablehead{CS\\Score} & \tablehead{PS\\Score} & \tablehead{Comp\\Score} & \tablehead{Case\\Acc.} \\
\midrule
\multirow{5}{*}{Qwen3.6-27B} & 0.1 & 86.1 & 82.8 & 84.4 & 31.2 \\
 & 0.3 & 86.0 & 82.1 & 84.1 & 30.0 \\
 & 0.5 & 86.7 & 82.5 & 84.6 & 32.1 \\
 & 0.7 & \textbf{86.9} & 83.8 & \textbf{85.3} & \textbf{33.8} \\
 & 0.9 & 85.6 & \textbf{84.6} & 85.1 & 32.5 \\
\midrule
\multirow{5}{*}{Qwen3.5-27B} & 0.1 & 77.7 & \textbf{81.7} & 79.7 & 18.6 \\
 & 0.3 & 79.0 & \textbf{81.7} & \textbf{80.3} & \textbf{19.6} \\
 & 0.5 & \textbf{79.4} & 77.5 & 78.5 & 19.0 \\
 & 0.7 & 78.2 & 79.6 & 78.9 & 17.9 \\
 & 0.9 & 76.5 & 77.6 & 77.1 & 17.0 \\
\midrule
\multirow{5}{*}{Sonnet 4.6} & 0.1 & 83.9 & \textbf{81.9} & 82.9 & 29.4 \\
 & 0.3 & 84.5 & 81.5 & \textbf{83.0} & 29.4 \\
 & 0.5 & \textbf{85.0} & 79.6 & 82.3 & 30.1 \\
 & 0.7 & 84.5 & 81.3 & 82.9 & \textbf{30.4} \\
 & 0.9 & 83.0 & 80.7 & 81.9 & 29.3 \\
\bottomrule
\end{tabular}
\end{table}

\clearpage
\subsection{SealQA Results}
\label{appendix:sealqa}

Table~\ref{tab:sealqa_full} reports the full metrics for the SealQA results in Table~\ref{tab:main_results} and Section~\ref{subsec:sealqa}, together with additional evaluations using Opus 4.8. C, NA, and W denote correct, not-attempted, and wrong answers, respectively, and sum to $N=111$. Accuracy is $100C/N$, and the partial-credit score (Mean) is $(C+0.5\,\mathrm{NA})/N$.

For HyperMCTS on SealQA, we use $\lambda=0.2$ for Qwen3.6-27B and Sonnet 4.6, and $\lambda=0.5$ for Qwen3.5-27B and Opus 4.8.

On Opus 4.8, HyperMCTS reaches $30.6\%$ accuracy, exceeding the strongest baselines, LATS and FLARE ($26.1\%$), by $4.5$ percentage points.

\begin{table}[!t]
\centering
\caption{Full SealQA results. Mean is on a $0$--$1$ scale; accuracy is in \%. \textbf{Bold} marks the best Mean and accuracy for each model, including ties.}
\label{tab:sealqa_full}
\setlength{\tabcolsep}{7pt}
\begin{tabular}{*{7}{c}}
\toprule
\multirow{2}{*}[-2.5pt]{Agent} & \multicolumn{3}{c}{Qwen3.6-27B} & \multicolumn{3}{c}{Qwen3.5-27B} \\
\cmidrule(lr){2-4} \cmidrule(lr){5-7}
 & Mean & Acc. (\%) & C/NA/W & Mean & Acc. (\%) & C/NA/W \\
\midrule
ReAct & 0.338 & 13.5 & 15/45/51 & 0.441 & 12.6 & 14/70/27 \\
\cmidrule[0.3pt](lr){1-7}
CoT & 0.324 & 14.4 & 16/40/55 & 0.437 & 13.5 & 15/67/29 \\
Reflexion & 0.311 & 14.4 & 16/37/58 & 0.401 & 13.5 & 15/59/37 \\
\cmidrule[0.3pt](lr){1-7}
ToT & 0.302 & 17.1 & 19/29/63 & 0.446 & \textbf{17.1} & 19/61/31 \\
RAP & 0.297 & 14.4 & 16/34/61 & 0.405 & 13.5 & 15/60/36 \\
LATS & 0.284 & 15.3 & 17/29/65 & 0.423 & 14.4 & 16/62/33 \\
FLARE & 0.347 & 18.9 & 21/35/55 & 0.432 & 14.4 & 16/64/31 \\
\midrule
\textbf{HyperMCTS} & \textbf{0.392} & \textbf{21.6} & 24/39/48 & \textbf{0.468} & \textbf{17.1} & 19/66/26 \\
\midrule
\multirow{2}{*}[-2.5pt]{Agent} & \multicolumn{3}{c}{Sonnet 4.6} & \multicolumn{3}{c}{Opus 4.8} \\
\cmidrule(lr){2-4} \cmidrule(lr){5-7}
 & Mean & Acc. (\%) & C/NA/W & Mean & Acc. (\%) & C/NA/W \\
\midrule
ReAct & 0.401 & 14.4 & 16/57/38 & 0.428 & 19.8 & 22/51/38 \\
\cmidrule[0.3pt](lr){1-7}
CoT & 0.369 & 14.4 & 16/50/45 & 0.428 & 20.7 & 23/49/39 \\
Reflexion & 0.374 & 16.2 & 18/47/46 & 0.405 & 22.5 & 25/40/46 \\
\cmidrule[0.3pt](lr){1-7}
ToT & 0.329 & 22.5 & 25/23/63 & 0.405 & 23.4 & 26/38/47 \\
RAP & 0.360 & 17.1 & 19/42/50 & 0.369 & 20.7 & 23/36/52 \\
LATS & 0.333 & 19.8 & 22/30/59 & 0.428 & 26.1 & 29/37/45 \\
FLARE & 0.329 & 15.3 & 17/39/55 & 0.423 & 26.1 & 29/36/46 \\
\midrule
\textbf{HyperMCTS} & \textbf{0.446} & \textbf{26.1} & 29/41/41 & \textbf{0.491} & \textbf{30.6} & 34/41/36 \\
\bottomrule
\end{tabular}
\end{table}

\end{document}

%% file: math_commands.tex
\usepackage{amsmath,amsfonts,bm}

\def\eqref#1{equation~\ref{#1}}

\def\1{\bm{1}}

\DeclareMathAlphabet{\mathsfit}{\encodingdefault}{\sfdefault}{m}{sl}
\SetMathAlphabet{\mathsfit}{bold}{\encodingdefault}{\sfdefault}{bx}{n}

